\documentclass{article}

\usepackage[preprint, nonatbib]{neurips_2021}

\usepackage[numbers]{natbib}

\usepackage[utf8]{inputenc} 
\usepackage[T1]{fontenc}    
\usepackage{hyperref}       
\usepackage{url}            
\usepackage{booktabs}       
\usepackage{amsfonts}       
\usepackage{nicefrac}       
\usepackage{microtype}      
\usepackage{xcolor}         

\usepackage{graphicx}

\usepackage[ruled]{algorithm2e}
\usepackage{multirow}

\usepackage{bbding}

\usepackage{subfigure}
\usepackage{float}

\title{Automating Neural Architecture Design without Search}

\author{%
  Zixuan Liang \\
  College of Computer Science\\
  Sichuan University, China\\
  \texttt{liangzixuan@stu.scu.edu.cn} \\
   \And
   Yanan Sun \\
   College of Computer Science \\
   Sichuan University, China \\
   \texttt{ysun@scu.edu.cn} \\
}

\begin{document}

\maketitle

\begin{abstract}

   Neural structure search (NAS), as the mainstream approach to automate deep neural architecture design, has achieved much success in recent years. However, the performance estimation component adhering to NAS is often prohibitively costly, which leads to the enormous computational demand. Though a large number of efforts have been dedicated to alleviating this pain point, no consensus has been made yet on which is optimal. In this paper, we study the automated architecture design from a new perspective that eliminates the need to sequentially evaluate each neural architecture generated during algorithm execution. Specifically, the proposed approach is built by learning the knowledge of high-level experts in designing state-of-the-art architectures, and then the new architecture is directly generated upon the knowledge learned. We implemented the proposed approach by using a graph neural network for link prediction and acquired the knowledge from NAS-Bench-101. Compared to existing peer competitors, we found a competitive network with minimal cost. In addition, we also utilized the learned knowledge from NAS-Bench-101 to automate architecture design in the DARTS search space, and achieved 97.82\% accuracy on CIFAR10, and 76.51\% top-1 accuracy on ImageNet consuming only $2\times10^{-4}$ GPU days. This also demonstrates the high transferability of the proposed approach, and can potentially lead to a new, more computationally efficient paradigm in this research direction.

\end{abstract}

\section{Introduction}

The impressive performance of a deep neural network depends largely on its architecture, which is usually designed by a handful of human experts with extensive knowledge in designing neural architectures. Unfortunately, this knowledge is not possessed by most interested users.  Therefore, how to automatically design a high-performance neural architecture has always been the interest of researchers. As the mainstream method, neural architecture search (NAS)\cite{elsken2019neural} formulate the architecture design as optimization problems, via solving these problems with well-designed optimization algorithms, the promising neural architectures are obtained\cite{DARTS, sun2020automatically, xie2018snas, sun2019completely}.

Affected by the optimization algorithms adopted, different NAS algorithms vary in the execution processes. Specifically, reinforcement learning (RL)\cite{kaelbling1996reinforcement} based NAS algorithms\cite{RL_NAS, zoph2016neural} use a controller to sample child networks, these child networks are trained to obtain the accuracies as rewards for controller. By maximizing these rewards, The controller learns to design satisfying neural architectures. Gradient-based NAS algorithms\cite{DARTS} relaxes the candidate operations with softmax to make the search space continuous, in order to use gradient descent to optimize the architecture with respect to its performance. Evolutionary computation (EC)\cite{back1997handbook}-based NAS algorithms\cite{real2019regularized, sun2019evolving} start with an initial population of neural architectures and generate better architectures iteratively by exploiting EC techniques. However, these NAS algorithms all have a common feature, they are constrained by the demands of enormous computation resources. 

NAS algorithms essentially consist of three parts: sampling, evaluating and updating. The time-consuming evaluation process is regarded by many researchers as the culprit for extensive computing resource requirements. Although various methods such as performance predictor\cite{sun2021novel, tang2020semi, liu2021homogeneous}, early stopping policy\cite{sun2018particle}  and  weight inheritance\cite{cai2018efficient, real2017large} have been proposed to speed up this process.  Regrettably, these methods do not solve the problem completely, because they failed to change the search-refine nature of the NAS. In addition, this search characteristic leads to another problem: the architecture design knowledge NAS learned during search is not transferable. When the search space changes, the distribution of network performance also changes. NAS algorithms need to be run from scratch again to obtain satisfactory results. This further causes the inefficiency of NAS.

To overcome the aforementioned limitations, we start from another direction to solve the problem of automatic neural architecture design. Different from those search methods, we learn from existing network structures and build a sort of (neural architecture, link guidance) knowledge repository which could provide link guidance based on the current network structure. By leveraging the feature extraction capability of graph neural networks (GNNs), we further introduce a GNN-based link prediction model to make discrete information in the knowledge repository continuous. In this way, the proposed method can design neural architecture in a computationally efficient way, and  could deal with unseen network structures even the search space changed. The contributions of this work could be summarized as follows:  

\begin{itemize}

\item We propose a novel approach to automatically design deep architectures dubbed as NAL(Neural Architecture Learning). Different from NAS approach that refine architectures through expensive performance evaluation, the proposed approach can directly achieve automation through learning knowledge of constructing high-performance deep architectures.

\item We implement the proposed approach through a GNN-based link prediction model. We further present an GNN upon the model to inversely aggregate node information for the node perception. This design could smooth the automation process.

\item We show that NAL can efficiently and effectively automate neural architecture for CIFAR-10 when the knowledge is obtained from NAS-Bench-101, and also be able to construct well-performing architecture for ImageNet when this knowledge is transferred to DARTS search space.

\end{itemize}

\section{Related Works}
This work is close to link prediction, cell-based neural architecture design, and representation learning on neural networks, which are briefly introduced in this section. 
\subsection{Link Prediction}

Link prediction is to predict whether an edge exists between two entities in a network. It is closely related  to various real-world application such as knowledge graph completion\cite{knowledgegraph1, knowledgegraph2} , recommend system\cite{recommender1, recommender2, recommender3}, and biological network reconstruction\cite{oyetunde2017boostgapfill}. As effective approaches for link prediction, Heuristic methods use various score functions to estimate the likelihood of links\cite{lu2011link}. Depending on the number of hop when calculating node's neighborhood, existing heuristics can be classified into first-order\cite{barabasi1999emergence}, second-order\cite{zhou2009predicting} and high-order\cite{jeh2002simrank}. \citet{lu2011link} has shown high-order heuristics performing better than first and second-order ones, and it has been proved by \cite{zhang2018link} that a large h is not necessarily needed to learn high-order graph structure features. As a matter of fact, these link prediction heuristics belong to graph structure features, which are located within the node and edge structures of the network. A number of algorithms have been proposed to extract the graph structure features, \citet{zhang2017weisfeiler} proposed Weisfeiler-Lehman neural machine to automatically learn suitable features from a network. They extract subgraph in nodes' neighborhood and train a neural network on these subgraph. For better graph feature extracting ability,  researchers began to focus on using GNN to handle this problem and achieving promising result\cite{zhang2018end, defferrard2016convolutional}. In this work we use Graph Attention Networks (GATs)\cite{velickovic2017graph} as our GNN model. With the help of attention mechanisms, GAT could highlight important information while reduce negative impact of noise in a automatic manner,which is exactly what we want when predicting the links in neural networks.

\subsection{Cell-based Neural Architecture Design}
Motivated by those well-performing neural networks\cite{he2016deep, huang2017densely} that have been artificially designed by stacking repeated motifs, \citet{zhong2018practical, RL_NAS} propose to form the final architecture by stacking the computation cells in a predefined way. Compared with layer-based and block-based neural architectures, this construction strategy makes the design of neural networks more dexterous and simple. Besides, it has been experimentally demonstrated by \cite{RL_NAS, szegedy2016rethinking, DARTS} that cell-based architectures could achieve superior performance and could be easily transferred to other datasets. Based on these advantages, we exploit the knowledge from existing computation cells and utilize it to design high-performance cell-based architectures. 

\subsection{Representation Learning on neural networks}
With the field of GNNs\cite{wu2020comprehensive, gori2005new, scarselli2008graph} which has seen a steep rise in interest, some inspiring researches\cite{dudziak2020brp} apply GNN to learn representation of neural networks. Specifically, 
\citet{li2020neural} combined GNN with variational autoencoder to map neural architecture to a continuous representation, then training a regression model to fit the performance of the continuous representation. \citet{knyazev2021parameter} leverages GNN to learn the representation of neural architectures and proposes a model that could directly predict the parameters of the input architectures. \citet{wen2020neural} uses Graph Convolutional Networks to generate feature embedding of the neural network architectures and pair these embeddings with their accuracies to train a neural predictor. They predict a large number of random architectures in the search space and fully train the top-K best predicted models, after which the best model is chosen as the final output. In contrast, we employ GAT to generate node representation in neural architectures, and predict beneficial link existence based on these node representations to design promising neural network.

\section{The proposed method}

The proposed NAL is a new perspective for automating neural architecture, the methodology is uniform and can be generalized to various Spaces. NAL constructs an experience database by exploiting the knowledge of existing network structure, based on these experiences, it could automatically design neural networks by predicting appropriate links. In the following section, we describe the details of NAL, which is composed of three parts: knowledge representation, experience storing, and structure prediction.

\subsection{knowledge representation}
\label{3.1}
\begin{figure}[htp]
    \centering
    \includegraphics[width=1\textwidth]{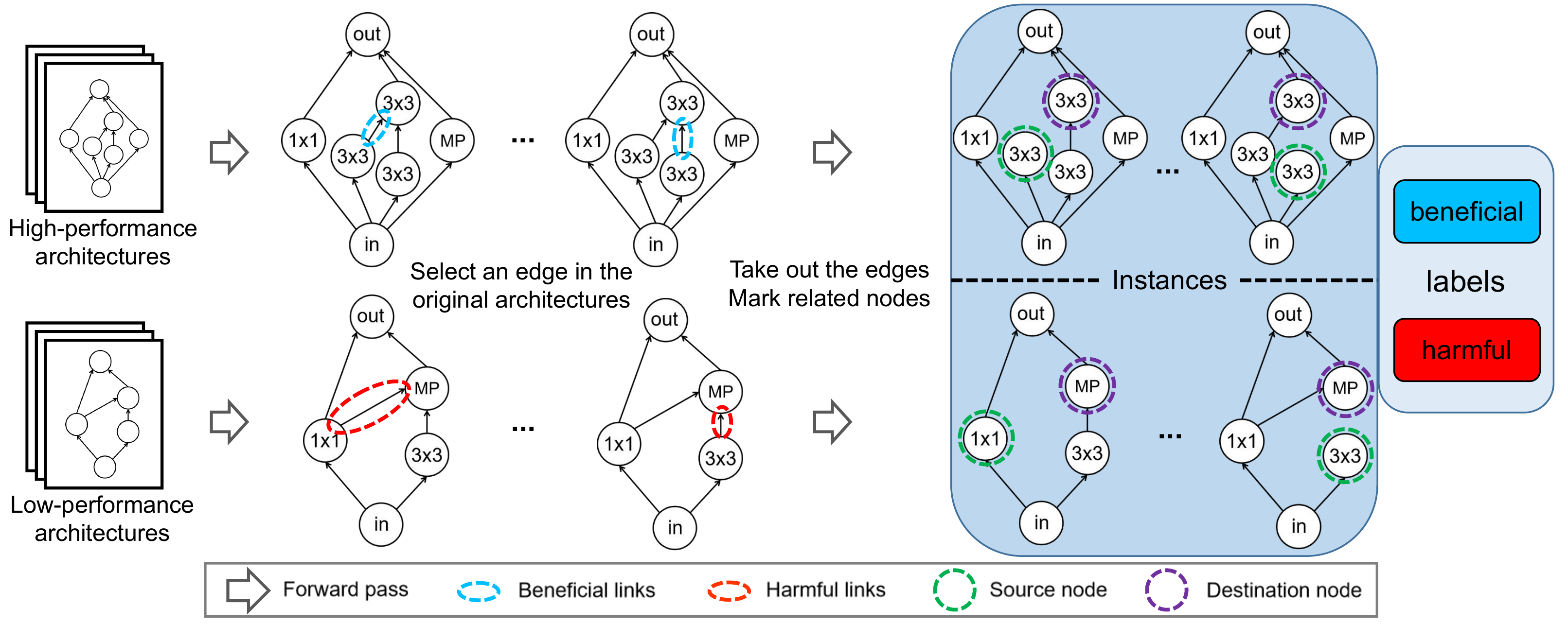}
    \caption{An illustration to the knowledge representation part of NAL. } 
    \label{fig1}
\end{figure}

The ultimate goal is to automat neural architecture design under the guidance of NAL, to achieve this goal, we need an appropriate way to express the knowledge of designing. Neural architecture is made up of operations and the links between them, In a given search space, the optional operations are fixed but the links between operations are uncertain. These links have a significant impact on the performance of the network, therefore, we decided to learn the design knowledge by studying links in existing networks. As shown in Figure \ref{fig1}, We select some high-performance and low-performance neural architectures to construct positive and negative training samples. Then randomly take out a link in the network architecture, and mark the source node and target node corresponding to this link. This marked incomplete network will be combined with its label to form a training sample.  These training samples will be used to train a GNN-based link prediction model, the model takes these incomplete network structures as inputs and predicts the links between the marked source nodes and target nodes. In order to make this link prediction model learn more rich design knowledge, we carefully designed the division of positive and negative samples for training data. For building positive samples,  we choose $K$ neural architectures in NasBench-101\cite{NASBench-101} with the highest test accuracy, each link in these architectures is taken out to form a  positive sample. In addition to this, we take apart the network structure step by step to mimic the network construction process, each time we remove an edge from the current network structure and form a  new sample, we randomly take a portion of them to join in the positive samples. For building negative samples, we take the same number $K$ of low accuracy network structures and form the negative samples in the same way we build positive samples. Furthermore, edges that do not exist in those highest test accuracy neural are also taken as negative samples. The final training data has consisted of the positive samples and the negative samples from the above process.

\subsection{Experience storing}
\begin{figure}[htp]
    \centering
    \includegraphics[width=1\textwidth]{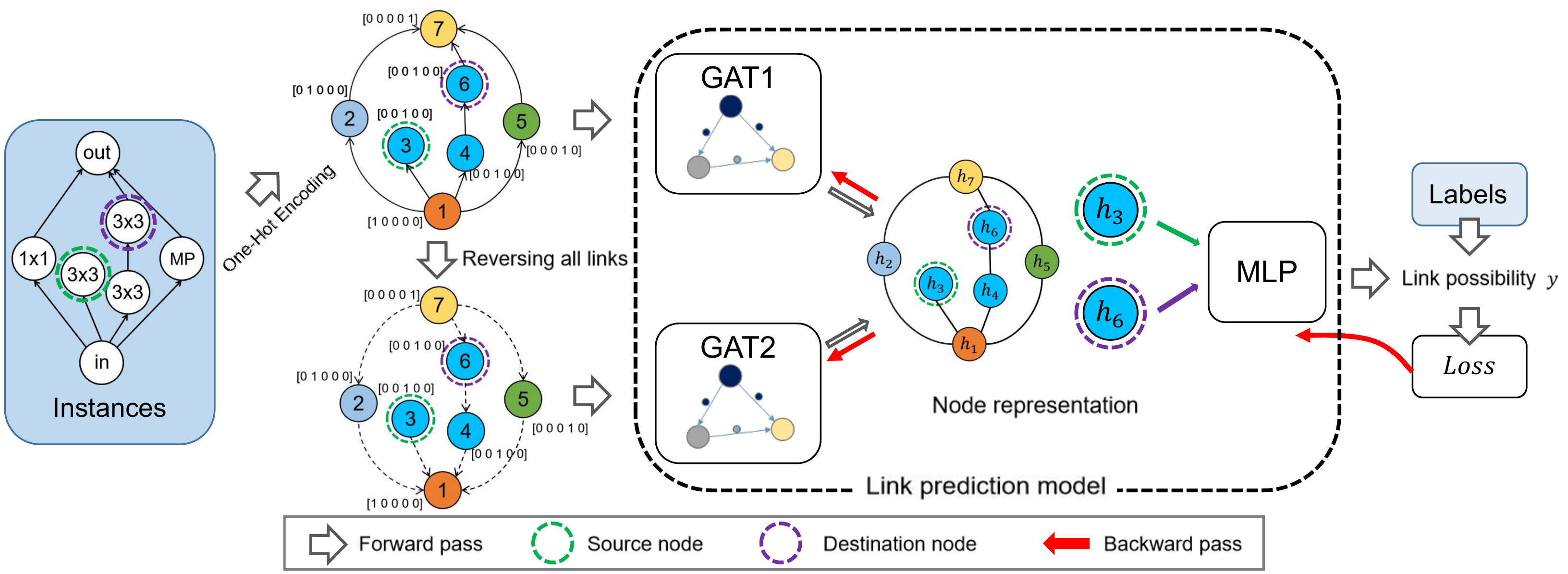}
    \caption{An illustration to the Experience storing part of NAL.} 
    \label{fig2}
\end{figure}
To learn and store experience from these training samples, we modify a GNN link prediction model to adapt to the task of link prediction in neural networks. As shown in Figure \ref{fig2}, a neural network architecture is represented by a DAG that uses nodes to represent operations and edges to represent the connections, the first step is to transform the representation of neural network architecture in order to facilitate the calculation of GNN model. We use the one-hot adjacency matrix encoding strategy which is commonly used in NAS researches, the $N$ nodes are encoded based on their corresponding operation by a one-hot vector $V =  \left\{V_1, V_2, ..., V_N \right\},  V\in \mathbb{R}^{N\times D} $, where $D$ is the number of possible operation types, and the connections of these nodes are contained in an adjacency matrix $A \in \mathbb{R}^{N\times N}$. Then the GNN model $\phi$ will compute representation $h$ for each node $i$ based on $V$ and $A$:
\begin{equation}
h_{i} = \phi (A, V)
\end{equation}
In neural networks, different operations have different meanings, their importance can not simply be equated. With this consideration, we choose the graph attention network model with three GAT layers as our GNN model. The input to GAT layer is a set of node embeddings $ h^{(l)} = \left\{ h_1^{(l)}, h_2^{(l)}, ..., h_N^{(l)}\right\}, h^{(l)}\in \mathbb{R}^F$ , where $l$ means current layer, $N$ is the number of nodes and $F$ is the number of feature in each node.  We initialize node features $h^{(0)}$ with their one-hot vector $V$ , each GAT layer $l$ computes the node embedding $h_{i}^{l}$  from the embedding of layer $l-1$ by leveraging the attention mechanism:
\begin{equation}
h_{i}^{(l)} = \sigma \Big(\sum_{j \in N_{(i)}} \alpha_{ij}^{(l-1)} {W}h_{j}^{(l-1)}  \Big)
\end{equation}
Where  $\sigma$ is an activation function,  $N{(i)}$ denotes the  one-hop neighbors of node $i$ in adjacency matrix $A$, $\alpha_{ij}^{(l-1)}$ indicates the attention coefficient of node $j$ to node $i$ in layer $l-1$, $W$ is a learnable matrix for node-wise feature transformation. Owing the links in neural architectures are unidirectional, which means the information can only flow in one direction when aggregating. This will lead to a problem that the source node is not aware of the information about those nodes linked by it, for example, the node embedding $h^{(l)}$of the input node will never change no matter how many nodes it has linked to. To solve this problem,  we add a feedback adjacency matrix $A^{\prime}$ by reversing all edges in the original graph. To avoid interference with the real links in the neural network, we introduce a separate GAT model  $\phi^{\prime}$ to compute embedding $h^{\prime}$ on the feedback adjacency matrix $A^{\prime}$ in the manner described above. By doing so, the embedding calculation of Equation 1 is redefined as:
\begin{equation}
\label{formulation 3}
h_{i} = \phi (A, V ) \parallel\phi^{\prime}(A^{\prime}, V)
\end{equation}
We concatenate the node embeddings calculated by these two GAT models as the final embedding $h_{i}$. The Basic idea of GNN based link prediction model is to calculate the score $y_{u, v}$ of link possibility between node u and v by their embedding. In practice, we use a MLP with two fully-connected layers to perform this calculation, the MLP takes node embedding $h_u$ and $h_v$ as input and outputs a one-dimensional vector that indicates the link possibility $y_{u, v}$. By iteratively minimizing  the cross-entropy loss between $y_{u, v}$ and label$t_{u, v}$ in Equation \ref{equation 4}:
\begin{equation}
\label{equation 4}
Loss = -(t_{u, v}\cdot \log(y_{u, v}) + (1 - t_{u, v})\cdot\log(1 - y_{u, v}))
\end{equation}
The link prediction model stores the experience of designing neural networks, and could clearly distinguish what links are beneficial and what links are detrimental to the network performance in terms of link possibility score $y_{u, v}$.

\subsection{Architecture prediction}
\begin{figure}[htp]
    \centering
    \includegraphics[width=1\textwidth]{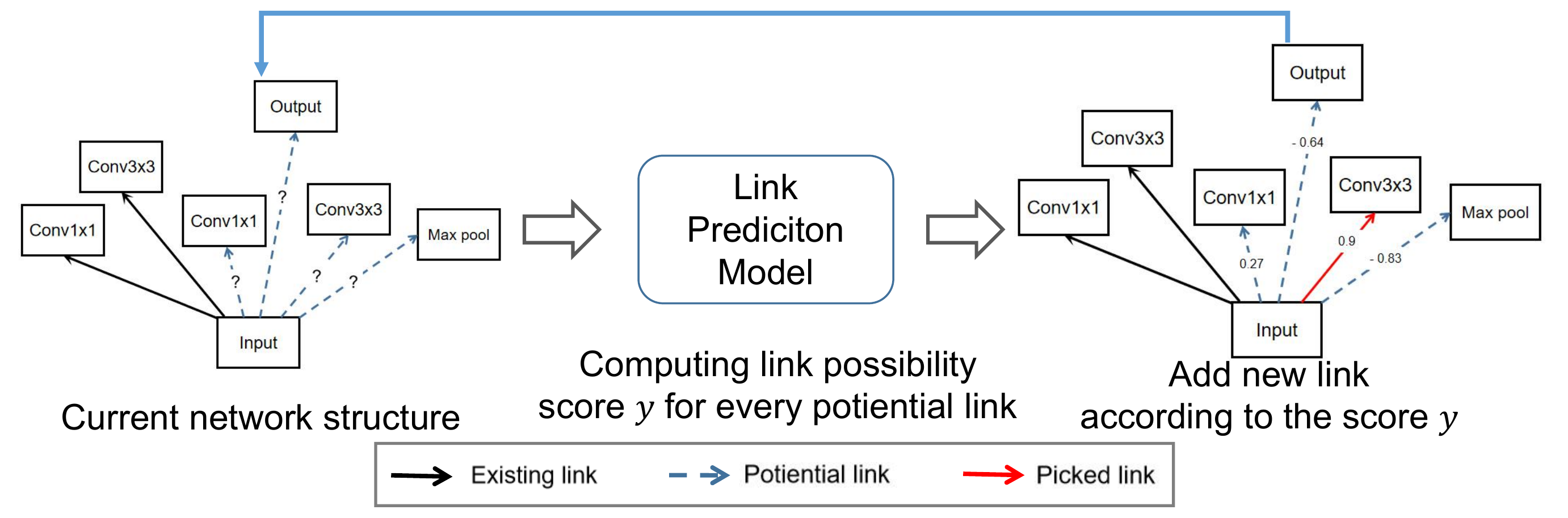}
    \caption{An illustration to the structure prediction part of NAL.} 
    \label{fig3}
\end{figure}

As shown in figure \ref{fig3}, the trained link prediction model is utilized to predict the score $y_{u, v}$ for all potential links in the current network structure. After that,  we add a new link based on their score to form a new network structure for the next round of prediction. This way of predicting new links based on the network structure allows the knowledge learned by NAL to be utilized in any search space. Algorithm \ref{algorithm1} shows the process of generating architectures using the link prediction model in the specified search space.

\begin{algorithm}[H]
\label{algorithm1}
\caption{Generate neural architecture with link prediction model}
\KwIn{The limitation of maximum edges $E$, the limitation of maximum vertexes $V$, the set of candidate operations $OP$.}
\KwOut{Architecture $\mathcal{A}$ generate by link prediction model.}
Initialize neural architecture $\mathcal{A}$ with a input node, initialize score pool $P$ as $\emptyset$. Current number of vertex $v$ and number of edge $e$ are initialized as 1 and 0. Source node set $Src$ and destiny node set $Dst$ are initialized as $\left\{input\right\}$ and $\emptyset$.

\While{$e < E$}{
\For{$i$ in $Src$}{
\For{$j$ in $Dst\parallel OP$}{
Calculate link possibility score $y_{i, j}$ between u and v by link prediction model\;
Add $y_{i, j}$ to $P$\;
}}
P = Softmax(P)\;
Sample a link $l_{u,v}$ from $P$ based on its value\ and add it to $\mathcal{A}$\;
Add v to $Src$ and $Dst$\;
$e = e + 1$\;
$v = v + 1$\;
\If{$v\geq V$}{
Remove $OP$\;
}
}\end{algorithm}

In the begining, the architecture is initialized with only a input node, the source node set $Src$ and destiny node set $Dst$ with candidate operations $OP$ are used to indicate the possible link in the network.  While current number of edge is not up to limit, we use link prediction model to give every possible edge a possibility score $y_{i,j}$ and add it to a score pool $P$. The node i is taken from $Src$ represents the starting point of link, the node j taken from $OP$ indicate the link between node i and a new node that not in current network while the node j taken from $Dst$ indicate the link between two nodes in current network. After regularized $P$ by the $Softmax$ function, we sample a link $l_{u, v}$ based on the score distribution from $P$, the link will be added to current architecture. Besides, the endpoint $v$ of the link is add to $Src$ and $Dst$ for the next round of predictions. In this way, we could get a generated network in the target search space.

\section{Experiments}
In this section, we carry out experiments on two search spaces ($i.e$, NASBench-101 and DARTS.) to demonstrate the effectiveness of the proposed NAL.

\textbf{NASBench-101}. NAS-Bench-101 is the first public architecture dataset for NAS research, which contain 423K unique cell-based convolutional architectures trained and evaluated multipule times on CIFAR-10. It allows researchers to directly query the performance of a large number of networks. Each cell in NAS-Bench-101 is consist of an input node, an output node and up tp five intermediate nodes, the limitation of maximum number of edges is 9. The nodes represent operations in neural network while the edge represent connection between operations. All intermediate nodes are selected from $3\times 3$ convolution, $1\times 1$ convolution or $3\times 3$ max-pool.

\textbf{DARTS}. As a popular search space for gradient-based NAS methods. The search space of DARTS is a computation cell represented by a DAG, different from NAS-Bench-101, each node is a latent representation while each edge is stand for operation. Every cell in DARTS is composed of 7 nodes, the first two nodes as input to receive the outputs from previous two cells respectively, the next four nodes as intermediate nodes with two input edge, the last 
node will concatenate all latent representation in the intermediate nodes as the output of current cell. The cell could be stacked to form a final architecture.

In the following, we first train  NAL on the neural architectures selected from NASBench-101, then introduce the experiment to demonstrate the capability of NAL in designing NASBench-101 networks. To further verify that NAL learned universal neural network design knowledge, we propose a search space mapping method that enables the NAL trained on NAS-Bench-101 to design neural architecture in DARTS search space.  We evaluate the performance of neural architecture designed by NAL with two classical datasets. In the end, we present the ablation experiment to demonstrate the impact of each part of the algorithm.

\subsection{Implementation Details}
To prepare the training data, we first sort all neural architectures in NAS-Bench-101 according to their test accuracy, then we choose 30 high-performance architectures and 30 low-performance architectures to build the training samples as mentioned in section \ref{3.1}. Due to the difference in the construction of positive and negative samples, the number of positive and negative samples is different. To avoid this difference affecting the network training, we take same number of samples randomly from the negative samples as the positive samples do. After one-hot encoding, the node representation in NAS-Bench-101 cell is a vector of dimension 5 because there are only five optional operations ($i.e.$ input, output, $3\times 3$ convolution, $1\times 1$ convolution and $3\times 3$ max-pool). The Graph Attention Network we used in NAL has three GAT layers, the input dimension is set to 5, the dimension of hidden layer and the final output are both 16, $Relu$ is used as the activation function. As mentioned in equation \ref{formulation 3}, we concatenate the embedding computed by two different GAT model as the final embedding of node, so the final embedding is a vector of dimension 32. The MLP we used in NAL consists of a fully-connected layer with hidden size 32 with $Relu$ and an output layer. For training details, we use the Adam optimizer\cite{kingma2014adam} with an initial learning rate 0.0005 and train  NAL for 200 epochs.

\subsection{Design Neural Architecture in NAS-Bench-101 search space}
\label{4.2}

\begin{table}[htp]
\caption{Performance comparison between NAL and prior work on Nas-Bench-101 dataset.}
\label{table1}
\centering
\begin{tabular}{cccc}

\hline
Methods       & Queries & Top-1 Accuracy(\%) & Strategy         \\ \hline
Random Search\cite{NASBench-101} & 1000    & 93.54             & Random                \\
RE\cite{NASBench-101}            & 1000    & 93.72             & Evolution             \\
RL\cite{NASBench-101}            & 1000    & 93.58             & REINFORCE             \\
BO\cite{NASBench-101}            & 1000    & 93.72             & Bayesian Optimization \\
NAO\cite{luo2018neural}           & 1000    & 93.81             & Gradient Decent       \\
E2EPP\cite{sun2019surrogate}         & 1000    & 93.77$\pm$0.13    & Evolution             \\
SSANA\cite{tang2020semi}         & 1000    & 94.01$\pm$0.12    & Evolution             \\ \hline
NAL           & \textbf{160}     & \textbf{94.03}$\pm$0.05    & Link Prediction       \\ \hline
\end{tabular}
\end{table}

We follow the design process mentioned in algorithm \ref{algorithm1} to design neural architectures in the NAS-Bench-101 search space.  The limitation of maximum edges E and maximum vertexes V in NAS-Bench-101 search space are set to 9 and 7 respectively, the set of candidate operations OP is \{input, $3\times 3$ convolution, $1\times 1$ convolution and $3\times 3$ max-pool, output\}. To assess the efficiency and quality of NAL, we referred to the NAS-Bench-101 experiments in \cite{tang2020semi, yan2020does}. To reduce the influence of random factors in the algorithm, we train 10 NAL models, each model designs 10 cell structures and we query their test accuracy in the NAS-Bench-101 dataset, the best one is selected as the final output of the proposed approach. Specifically, the experiment is repeated 30 times, Table \ref{table1} shows the comparison result, two evaluation indicators are used to evaluate the efficiency and quality of the algorithm respectively: the number of architecture quires and final test accuracy on the CIFAR-10 dataset. Appending the 60 training architectures mentioned before, the total queries number is 160, which means that NAL could design a more accurate architecture while its efficiency is significantly better than other methods. 

\begin{figure}[htp]
    \centering
    \includegraphics[width=1\textwidth]{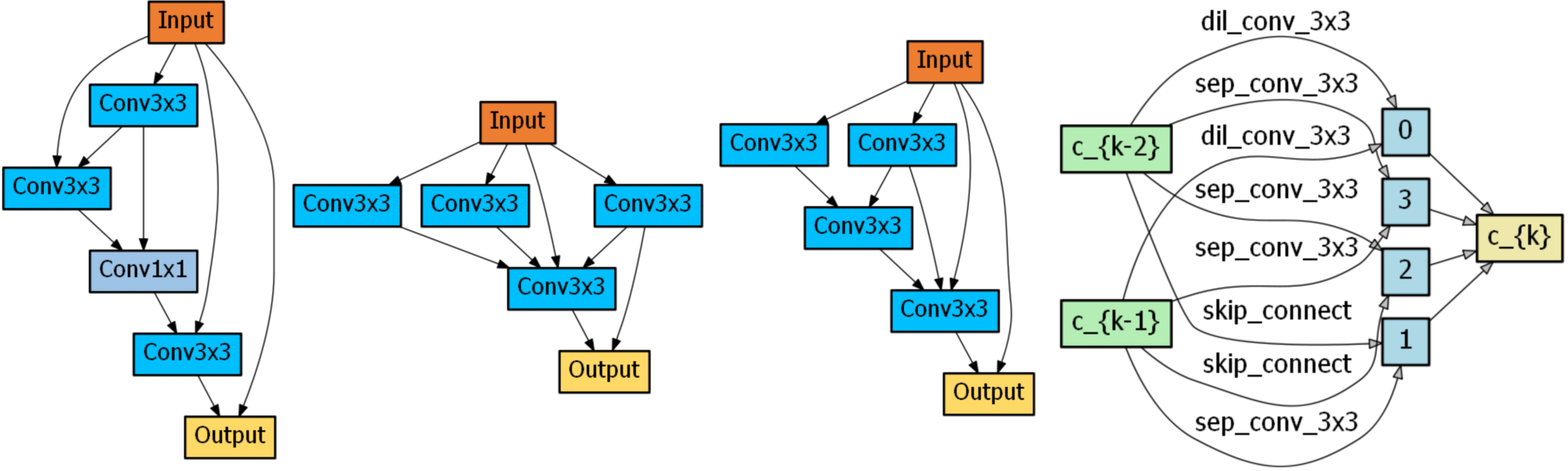}
    \caption{The cells designed by NAL in NAS-Bench-101 and DARTS search space.}
    \label{fig4}
\end{figure}


Figure \ref{fig4} shows some of the cell structures that frequently appear in NAL design results, when designing neural architecture in NAS-Bench-101, NAL hardly selects edges associated with $3\times 3$ max-pool; on the contrary, it is more inclined to select $3\times 3$ convolution as the operation of intermediate nodes or to combine $1\times 1$ convolution with $3\times 3$ convolution. Furthermore, NAL tends to link a large number of intermediate nodes for input node to increase the width of the network at the front.

\subsection{Design Neural Architecture in DARTS search space}


As mentioned above, the representation of cell and the operations in DARTS are quite different from those operations in NAS-Bench-101. In order to apply the trained NAL to DARTS search space, we need additional transformations to get a adjacency matrix $A$ and a node feature list $V$ and map the operations in DARTS to the operations in NAS-Bench-101. We replace every intermediate node that has two input edges in DARTS with two nodes that have the same index,  the two new nodes represent the operation of two original input edges respectively. For those operations in DARTS that do not appear in NAS-Bench-101, We combine the operations in NAS-Bench-101 to form local structures that have the same receptive field as alternatives. The detail of the transformation process and operation replacement strategy could be seen in supplementary materials.



After transformation, we apply the NAL trained on NAS-Bench-101 to design architecture in the DARTS search space, the neural architectures designed by NAL are shown in Figure \ref{fig4}. When designing neural architectures in DARTS search space, NAL is inclined to choose  $3 \times 3$ dilated separable convolutions and $3 \times 3$ separable convolutions. Besides, NAL focuses more on expanding the network's width than its depth, this is similar to the design characteristic of NAL in the NAS-Bench-101 search space. In the following section, we evaluate the architecture designed by NAL with CIFAR-10 and ImageNet\cite{deng2009imagenet} dataset.

\subsection{Results on CIFAR-10 Image Classification}

\begin{table}[htp]
\caption{Performance comparison on CIFAR-10 dataset with state-of-the-art methods.}
\label{table2}
\centering
\begin{tabular}{ccccc}
\toprule
Methods     & Cost (GPU-day) & Test Error(\%) & Params(M)  & Strategy \\ \midrule
NASNet-A\cite{RL_NAS}    & 1,800      & 2.65       & 3.3               & RL            \\
ENAS\cite{pham2018efficient}        & 0.5       & 2.89       & 4.6                & RL            \\
AmoebaNet-B\cite{real2019regularized} & 3,150     & 2.55$\pm$0.05  & 2.8              & evolution     \\ 
DARTS\cite{DARTS}       & 4         & 2.76$\pm$0.09  & 3.3                  & gradient      \\
SNAS\cite{xie2018snas}        & 1.5        & 2.85$\pm$0.02  & 2.8               & gradient      \\ 
P-DARTS\cite{chen2019progressive}     & 0.3       & 2.50       & 3.4                & gradient      \\
PC-DARTS\cite{xu2019pc}    & 0.1       & 2.57$\pm$0.07  & 3.6                & gradient      \\
PR-DARTS\cite{zhou2020theory}    & 0.17      & 2.32       & 3.4               & gradient      \\
Fair DARTS\cite{chu2020fair}  & 0.4       & 2.54       & 2.8                & gradient      \\
ISTA-NAS\cite{yang2020ista}\  & 2.3       & 2.36$\pm$0.06     & 3.37                & gradient      \\
\midrule
NAL        & \textbf{2e-4}      & \textbf{2.18}       & 3.2               & Link Prediciton\\\bottomrule

\end{tabular}
\end{table}
Following the setting of \cite{DARTS}, our network is formed by stacking 20 cells with 36 initial channels. We use SGD to optimize parameters with the initial learning rate of 0.025 scheduled by Cosine Annealing LR. The CIFAR-10 results are present in Tabel \ref{table2}. Notably, the proposed NAL achieves a state-of-the-art error rate of 2.18\% with the size of 3.2M, the training of NAL only takes 2e-4 GPU-day (around 20 seconds).   
\subsection{Results on ImageNet Image Classification}

\begin{table}[htp]
\caption{Performance comparison on ImageNet dataset with state-of-the-art methods.}
\label{table3}
\centering
\begin{tabular}{cccccc}
\toprule
\multirow{2}{*}{Methods} & Cost & \multicolumn{2}{l}{Test Error(\%)} & Params          & \multirow{2}{*}{\begin{tabular}[c]{@{}l@{}}Strategy \\\end{tabular}} \\
                         & (GPU-Day) & Top-1              & Top-5         & (M)        &                                                                           \\
\midrule
Inception-V1\cite{szegedy2016rethinking}             & -         & 30.2               & 10.1          & 6.6     & manual\\
MobileNet\cite{howard2017mobilenets}                & -         & 29.4               & 10.5          & 4.2     & manual\\                                                                    
ShuffleNet v2 2x\cite{ma2018shufflenet}         & 0.5 & 25.1               & -             & $\sim$4.6        & RL\\                                                \midrule                        
NASNet-A\cite{RL_NAS}                 & 1,800     & 26.0               & 8.4           & 5.3     & RL\\
AmoebaNet-C\cite{real2019regularized}              & 3,150     & 24.3               & 7.6           & 6.4     & evolution\\
MnasNet-92\cite{tan2019mnasnet}               & -         & 25.2               & 8.0           & 4.4     & RL\\
ISTA-NAS\cite{yang2020ista}                & 33.6      & 24.0               & 7.1           & 5.65    & gradient\\
DARTS\cite{DARTS}    & 4         & 26.7               & 8.7           & 4.7     & gradient\\
SNAS\cite{xie2018snas}              & 1.5       & 27.3               & 9.2           & 4.3     & gradient\\
P-DARTS\cite{chen2019progressive}                  & 0.3       & 24.4               & 7.4           & 4.9     & gradient\\
BayesNAS\cite{zhou2019bayesnas}                 & 0.18      & 26.5               & 8.9           & 3.9     & gradient\\
PC-DARTS\cite{xu2019pc}                 & 0.13      & 25.1               & 7.8           & 5.3     & gradient\\
GDAS\cite{dong2019searching}                     & 0.21      & 26.0               & 8.5           & 5.3     & gradient\\
Fair DARTS\cite{chu2020fair}               & 0.4       & 24.9               & 7.5           & 4.8     & gradient\\
PR-DARTS\cite{zhou2020theory}                 & 0.17      & 24.1               & 7.3           & 4.98    & gradient\\
\midrule
NAL                     & \textbf{2e-4}      &\textbf{23.5}                & \textbf{7.0}           & 5.0       & link prediction\\
\bottomrule          
\end{tabular}
\end{table}
For ImageNet image classification, the final network is formed by stacking 14 cells designed by NAL and 48 initial channels. The network is trained from scratch for 350 epochs with batch size 512 on 4 RTX 2080Ti GPUs. We use SGD to optimize parameters with the initial learning rate of 0.2 scheduled by Cosine Annealing LR, the momentum and weight decay are set to 0.9 and $4\times 10^{-4}$. Experiment results on ImageNet are present in Table \ref{table3}. Architecture designed by NAL outperforms current cell-based NAS methods with comparable model parameters while having minimal search cost.
\subsection{Ablation Study}
\begin{table}[htp]
\caption{Comparison on NAS-Bench-101. S, I and G are short for "Sample construction strategy", "Inversely information aggregating" and "GAT" seprately. The italic and bold denote the best values of each evaluation indicators}
\label{table4}
\centering
\begin{tabular}{ccc|cc}
\hline
S & I & G & Mean Accuracy(\%) & Top-1 Accuracy(\%) \\ \hline
\XSolidBrush & \XSolidBrush & \Checkmark & 92.58$\pm$0.05               & 93.87$\pm$0.09                 \\
\XSolidBrush & \Checkmark & \Checkmark & 92.77$\pm$0.06               & 93.93$\pm$0.09                \\
\Checkmark & \XSolidBrush & \Checkmark & 93.36$\pm$0.08    & 93.91$\pm$0.04      \\
\Checkmark & \Checkmark & \XSolidBrush & 93.44$\pm$0.05    & 93.98$\pm$0.08      \\ \hline
\Checkmark & \Checkmark & \Checkmark & \textbf{93.60}$\pm$0.04    & \textbf{94.03}$\pm$0.05        \\ \hline
\end{tabular}
\end{table}
In this subsection, we carry out the ablation study in the NAS-Bench-101 search space for convenience. There are two evaluation indicators, the average accuracy of the designed 100 architectures mentioned in section \ref{4.2} that indicate the stability of the algorithm. And the top-1 accuracy as the final output of NAL indicates the quality of the algorithms.

\textbf{The impact of training sample construction strategy.} As mentioned in subsection \ref{3.1}, we carefully designed the positive and negative samples to make the GNN-based link prediction model adapt to the task of link prediction in neural networks. To verify the effect of this construction strategy on NAL, we compare it with the common practice of training sample construction in the link prediction task. The common practice only takes out links from unabridged high-performance architectures to form positive sample and only take the edges that do not exist in these architectures as negative samples. As can be seen from Table \ref{table4}, there are significant differences in the performance of NAL between these two sample construction strategies. With the help of network disassembly process and the Negative sample provided by low-performance networks, NAL can learn the design knowledge of neural network more stably.

\textbf{The impact of inversely information aggregating.} To solve the node perception problem, we reverse the original links and use an extra GAT to aggregate node information. For comparison, we disabled this additional GAT with other Settings unchanged, Table 4 shows the difference between these two sample construction strategies, when the inversely information aggregating is banned, nodes in the network cannot perceive their outbound links. This will affect the sensitivity of NAL to the change of network structure, thus, the performance of the algorithm is impaired.

\textbf{The impact of GAT.} Compare with GCN, which is the common practice of neural predictors, GAT gives different weights to different neighbors when aggregating neighbor information, whereas GCN treats each neighbor as equally important. We tested the effect of these two GNN models on NAL under the same setting. As shown in Table 4, when the GNN model in NAL changes from GAT to GCN, the stability and quality of NAL are slightly reduced. This difference validates the viewpoint that different nodes in neural networks have different contributions to the overall performance, and their importance cannot be equalized easily.

\section{Conclusion}
In this paper, we provide a new approach for automatic design of neural networks, by exploiting the knowledge of existing network structure, NAL could design high performance neural architecture by predicting new links for current network. The proposed approach could relieve the enormous requirements of extensive computing resources for automatic design neural network, and make the design process of neural architecture transparent to users. We have demonstrate its effectiveness in design neural architectures with promising performance in different search space, and further show that the knowledge NAL learned from one search space could be transfer to other search space. In the future, we plan to explore the potential of NAL to exploit the knowledge of existing network structure in a more efficient and stable way.


\bibliographystyle{plainnat}
\bibliography{ref}
\end{document}